\documentclass{article}

 \usepackage[preprint]{neurips_2026}

\usepackage[utf8]{inputenc} 
\usepackage[T1]{fontenc}    
\usepackage{url}            
\usepackage{booktabs}       
\usepackage{amsfonts}       
\usepackage{nicefrac}       
\usepackage{microtype}      
\usepackage{xcolor}         

\usepackage{amsmath}
\usepackage{graphicx} 
\usepackage{subcaption}
\usepackage{booktabs}
\usepackage{multirow}
\usepackage[table,xcdraw]{xcolor}
\usepackage{array}
\usepackage{makecell}
\usepackage{wrapfig}
\usepackage{enumitem}

\definecolor{mydarkblue}{rgb}{0,0.08,0.45}
\usepackage[colorlinks,citecolor=mydarkblue,urlcolor=mydarkblue,linkcolor=mydarkblue]{hyperref}
\usepackage{cleveref}      

\usepackage{pifont}
\newcommand{\cmark}{\ding{51}}
\newcommand{\xmark}{\ding{55}}

\usepackage{caption}  

\title{Q-ARVD: \\Quantizing Autoregressive Video Diffusion Models}

\author{%
  Siao Tang$^1$ \quad
  Xinyin Ma$^1$ \quad
  Gongfan Fang$^1$ \quad
  Xingyi Yang$^2$ \quad
  Xinchao Wang$^1$ \thanks{Corresponding author} \\[0.4em]
  $^1$National University of Singapore \quad
  $^2$The Hong Kong Polytechnic University \\[0.3em]
  \texttt{\{siao, maxinyin, gongfan\}@u.nus.edu} \quad
  \texttt{xingyi.yang@polyu.edu.hk} \quad \\[0.3em]
  \texttt{xinchao@nus.edu.sg}
  \vspace{-2em}
}

\begin{document}

\maketitle

\begin{abstract}
Autoregressive video diffusion models (ARVDs) have emerged as a promising architecture for streaming video generation, paving the way for real-time interactive video generation and world modeling. Despite their potential, the substantial inference cost of ARVDs remains a major obstacle to practical deployment, making model quantization a natural direction for improving efficiency. However, quantization for ARVDs remains largely unexplored. Our empirical analysis shows that directly applying existing quantization schemes developed for standard diffusion transformers to ARVDs leads to suboptimal performance, revealing quantization behaviors that differ from those observed in bidirectional diffusion models. In this paper, we identify two critical challenges in quantizing ARVDs: \textbf{(C1)} Highly unbalanced frame-wise quantization sensitivity. Error accumulation during autoregressive generation can induce severely skewed quantization sensitivity across frames, following an exponential-like decay pattern. \textbf{(C2)} Prominent and heterogeneous outlier patterns in weights. Weight distributions exhibit pronounced outlier channels, whose patterns vary substantially across layer types and block depths. To address these issues, we propose \textbf{Q-ARVD}, a novel framework for accurate ARVD quantization. (S1) To tackle the highly unbalanced frame-wise sensitivity, Q-ARVD incorporates a final-quality aware frame-weighting mechanism into the quantization objective. (S2) To prevent heterogeneous outliers from degrading performance, Q-ARVD introduces an outlier-aware adaptive dual-scale quantization, which automatically detects the presence and quantity of outlier channels for an arbitrary layer, and isolates them to protect normal channels. Extensive experiments on state-of-the-art open-source ARVDs (i.e., self-forcing and causal-forcing) demonstrate the superiority of Q-ARVD. Practical deployment of INT8 model shows 1.30x speedup and 1.97x model size reduction. \href{https://github.com/tsa18/Q-ARVD}{Code available here}.
\end{abstract}
\section{Introduction}

Video diffusion models~\citep{wan2025wan,hacohen2024ltx,kong2024hunyuanvideo,yangcogvideox,wu2025hunyuanvideo} have demonstrated strong capabilities in high-fidelity and temporally coherent video content generation. While traditional bidirectional video diffusion models excel at offline generation, they fundamentally struggle with real-time interactive applications due to their full-sequence joint generation paradigm. Recently, Autoregressive Video Diffusion Models (ARVDs)~\citep{huang2025selfforcing,zhu2026causal,yin2025slow,teng2025magi,jin2024pyramidal,chen2025skyreels,dengautoregressive} have emerged as an appropriate architecture for streaming video generation. By transforming video synthesis into a chunk-by-chunk or frame-by-frame causal generation process, ARVDs pave the way for applications such as real-time interactive video content generation~\citep{shin2025motionstream,ki2026avatar,feng2025streamdiffusionv2} and world modeling~\citep{mao2025yume,sun2025worldplay,huang2025selfforcing}.

Similar to other foundation models, enhancing the inference efficiency of ARVDs by model quantization~\citep{nagel2021white,krishnamoorthi2018quantizing} is of great practical importance, particularly for real-time scenarios and deployment on resource-constrained devices.
However, directly applying quantization to ARVDs remains a non-trivial endeavor due to the paradigm shift from bidirectional to autoregressive. Off-the-shelf quantization methods optimized for bidirectional diffusion transformers~\citep{wu2024ptq4dit,lisvdquant} or large language models (LLMs)~\citep{xiao2023smoothquant} often yield suboptimal performance. In this work, we bridge this gap by identifying and addressing two bottlenecks that uniquely characterize the quantization of ARVDs.

\textbf{First}, we observe a \textit{highly unbalanced quantization sensitivity across frames} caused by error accumulation. In ARVDs, the generation of the current frame is conditioned on the past generated frames. Consequently, quantization errors introduced in early frames rapidly compound over the autoregressive rollout. This suggests that frame-wise quantization sensitivity is heavily skewed toward early frames. Our empirical study reveals that this sensitivity follows an exponential-like decay along the temporal axis, indicating that the quality of the generated video is disproportionately governed by the precision of the early frames. As a result, treating all frames equally during quantization calibration is sub-optimal. \textbf{Second}, we observe that weight distributions in ARVDs exhibit \textit{prominent channel-wise outliers}. A small fraction of input channels (e.g., 2.1\%) show substantially larger magnitudes than the majority, elevating the difficulty of quantization. Furthermore, these outlier patterns are highly heterogeneous, varying markedly across different layer types (e.g., self-attention, cross-attention, and FFN) and block depths. Some layers exhibit severe outliers, while certain layers are well-behaved, so a static solution for outliers is inherently not appropriate.

To tackle the two challenges, we propose \textbf{Q-ARVD}, a novel quantization framework specifically tailored for autoregressive video diffusion models. To cope with the first challenge, i.e., the unbalanced frame-wise sensitivity, Q-ARVD introduces a \textit{final-quality guided frame-weighting} mechanism into the quantization objective. 
We directly quantify this sensitivity by evaluating how quantizing a certain frame affects the overall generated video, thereby modeling the actual effect of autoregressive error propagation.
We then assign importance weights to different frames during quantization calibration, emphasizing precision preservation for critical early frames.
To address the second challenge, i.e., the heterogeneous outlier patterns, Q-ARVD proposes an \textit{outlier-aware adaptive dual-scale} quantization. This strategy automatically identifies the presence and optimal number of outlier channels for arbitrary layers. To prevent the identified outlier channels from interfering with normal channels, we employ separate quantizers for them, resulting in a lower scaling factor for normal channels and thereby reducing quantization errors. Our main contributions can be summarized as follows:

\begin{itemize}[leftmargin=2em]
    \item We identify two critical challenges for quantizing autoregressive video diffusion models, i.e., unbalanced frame-wise quantization sensitivity, and prominent heterogeneous outlier patterns of model weights.
    \item To resolve the two challenges, we propose \textbf{Q-ARVD}, which features a final-quality guided frame-weighting mechanism to handle sensitivity discrepancy, and an adaptive dual-scale strategy to automatically detect and address outliers. To the best of our knowledge, Q-ARVD is the first quantization framework tailored for autoregressive video diffusion models.  
    \item Extensive experiments demonstrate that Q-ARVD significantly outperforms existing diffusion quantization baselines, achieving near-lossless visual quality.
    In practical deployment, the INT8 model delivers a $1.97\times$ reduction in model size and a $1.30\times$ latency speedup.
\end{itemize}

\section{Related Works}

\subsection{Autoregressive Video Diffusion Models}
Recent video generation models are shifting from full-sequence bidirectional generation~\citep{wan2025wan,kong2024hunyuanvideo,yangcogvideox} to autoregressive generation~\citep{teng2025magi,huang2025selfforcing,zhu2026causal}. Similar to causal decoding in large language models, autoregressive video diffusion models generate frames or chunks sequentially, conditioning each new frame on previously generated ones, 
formulated as $P(\boldsymbol{x}_0^{1:N}) = \prod_{i=1}^N p_\theta(\boldsymbol{x}_0^i \mid \boldsymbol{x}_0^{<i}),$ where $p_\theta(\boldsymbol{x}_0^i \mid \boldsymbol{x}_0^{<i})$ is modeled by diffusion denoising conditioned on past clean frames.
Early ARVDs rely on multi-step diffusion denoising and thus suffer from high inference latency. Recent methods improve efficiency and quality through few-step distillation, exposure-bias mitigation, and teacher-student architecture alignment~\citep{yin2025slow,huang2025selfforcing,zhu2026causal}, while another line of work extends fixed-length autoregressive models to long-horizon generation~\citep{yang2025longlive,yesiltepe2025infinity,liu2025rolling,yi2025deep}. These advances make ARVDs well-suited for streaming video generation, enabling real-time interactive generation~\citep{shin2025motionstream} and world modeling~\citep{sun2025worldplay}. Our work further improves their inference efficiency, particularly for deployment on resource-constrained devices.

\subsection{Model Quantization Preliminaries}

Model quantization~\citep{nagel2021white,krishnamoorthi2018quantizing} is one of the most significant techniques of efficient model inference. Quantization methods compress neural networks by representing model weights and input activations using low-precision formats, e.g., INT4 or INT8. The quantization process can be formulated as:

{
\setlength{\abovedisplayskip}{-1em}
\setlength{\belowdisplayskip}{-0em}
\begin{align}
x_q= \operatorname{clip}\left(\operatorname{round}\left(\frac{x}{s}\right)+z, q_{\min }, q_{\max }\right),
\label{eq:quantization}
\end{align}
}

\noindent where $x_q$ is the low-precision representation, $s$ is the scaling factor, and $z$ is the zero-point. For symmetric quantization, $s=\frac{\max(|x|)}{2^{b-1}-1}$. $q_{\min } \ \text{and} \ q_{\max }$ denote the lower and upper bounds of the low-precision format. To better maintain the model performance, a common practice~\citep{nagel2020up,librecq} is to optimize quantization parameters through reconstruction on a calibration dataset $\mathcal{D}_{cal}$:

{
\setlength{\abovedisplayskip}{-1.3em}
\setlength{\belowdisplayskip}{0.4em}
\begin{align}
\mathcal{L} = \mathbb{E}_{\mathbf{X} \sim \mathcal{D}_{cal}} \left\| \mathbf{X} \mathbf{W} - Q(\mathbf{X}) Q(\mathbf{W}) \right\|_\text{F}^2,
\label{eq:recon}
\end{align}
}

where $\mathbf{X}$ and $\mathbf{W}$ are activations and weights. $Q(.)$ denotes the quantize-then-dequantize operation. Learnable parameters include scaling factors, rounding schemes, etc.

\subsection{Model Quantization for Diffusion Models}

Quantization methods have been widely applied to improve the inference efficiency of diffusion models. Early works~\citep{shang2023post,li2023q,he2023ptqd,so2023temporal,huang2024tfmq,tang2024post} focus on quantizing the UNet backbone in diffusion models, and incorporate specific designs to accommodate the temporal denoising characteristics. With the architectural shift toward diffusion transformers (DiTs)~\citep{peebles2023scalable}, subsequent works~\citep{wu2024ptq4dit,lisvdquant,zhaovidit,li2025dvd,feng2025q,huang2025qvgen} propose dedicated quantization schemes designed for DiT-based diffusion models. 
Likewise, the recent paradigm shift from bidirectional to autoregressive video diffusion introduces new challenges for quantization, as mentioned before. Motivated by this, we develop a quantization framework tailored for autoregressive video diffusion models.

\section{Method: Q-ARVD}

\begin{figure*}[htbp]
    \centering
    \includegraphics[width=0.99\linewidth]{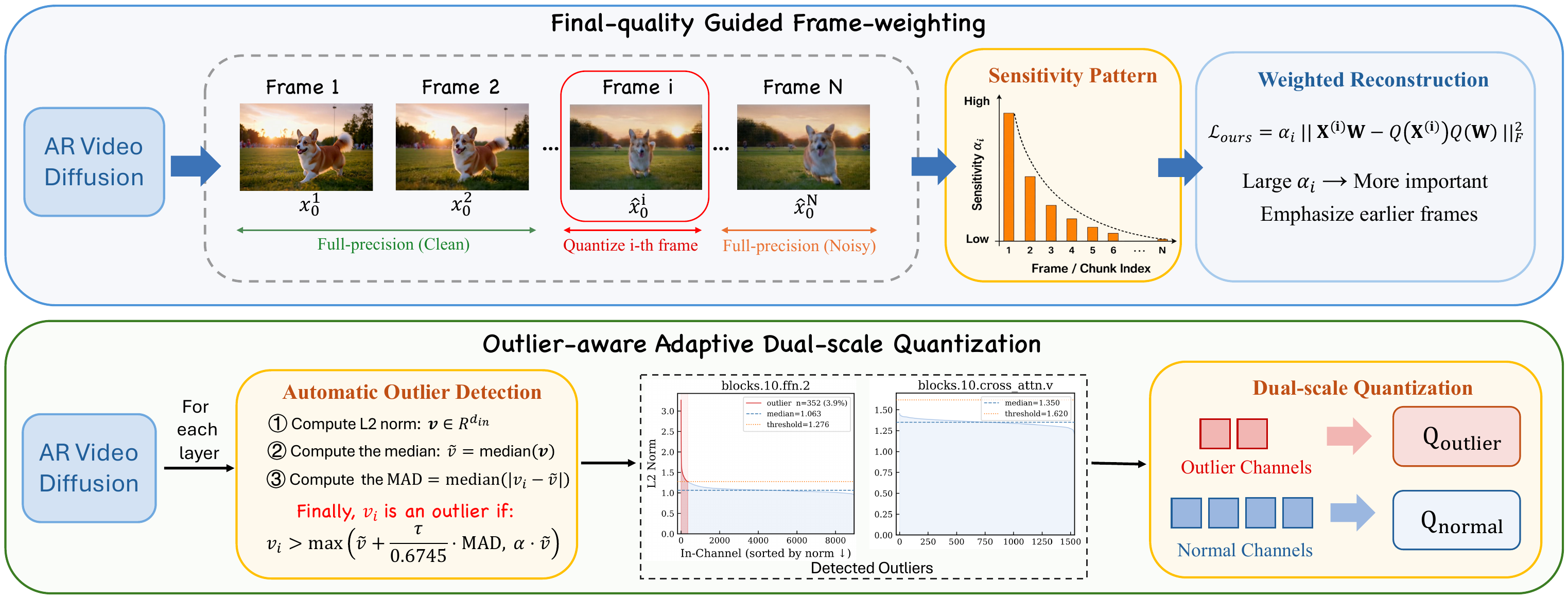}
    \caption{The illustration of our Q-ARVD framework.}
    \label{fig:framework}
    \vspace{-1.5em}
\end{figure*}

In this section, we elaborate on the proposed Q-ARVD framework. There are two key innovations of our framework. First, to address the issue of unbalanced frame-wise sensitivity, we propose the \textit{final-quality guided frame-weighting} mechanism (\S\ref{sec:method-weighting}). Second, to deal with the heterogeneous outlier patterns in model weights, we introduce an \textit{outlier-aware adaptive dual-scale} quantization strategy (\S\ref{sec:dual_scale}). The overall framework is illustrated in ~\Cref{fig:framework}.

\subsection{Final-quality Guided Frame-weighting}
\label{sec:method-weighting}

Autoregressive video diffusion models combine high-quality diffusion sampling with the LLM-like autoregressive decoding paradigm. The generation of a new frame $\boldsymbol{x}_0^{i}$ is conditioned on previous clean frames $\boldsymbol{x}_0^{<i}$. However, unlike discrete tokens in LLMs, these frames are continuous and high in information density, making the errors in previous frames significantly undermine subsequent frames. Intuitively, earlier frames exert a greater impact on the overall video quality.
In other words, the quality of generated videos is more sensitive to the quantization errors in earlier frames. 

We formally denote the \textit{frame-wise sensitivity} of the $i$-th frame as $\alpha_i$, where a larger $\alpha_i$ indicates higher sensitivity. To accurately quantify $\alpha_i$, we employ the final video quality degradation as a direct indicator, which we find is simple but effective.
Specifically, let $V=\{\boldsymbol{x}_0^{i}\}_{i=1}^{N}$ denote a video with $N$ frames, where $\boldsymbol{x}_0^{i}$ is the $i$-th clean frame.
The original autoregressive generation is
$P(\boldsymbol{x}_0^{1:N}) = \prod_{i=1}^N p_\theta(\boldsymbol{x}_0^i \mid \boldsymbol{x}_0^{<i}).$ To calculate $\alpha_i$, we only enable quantization for the generation of the $i$-th frame. Then, the modified autoregressive process is:

{
\setlength{\abovedisplayskip}{-1.2em}
\setlength{\belowdisplayskip}{0em}
\begin{align}
\hat{P}_i(\hat{\boldsymbol{x}}_0^{1:N}) = \underbrace{\prod_{k=1}^{i-1} p_\theta(\boldsymbol{x}_0^k \mid \boldsymbol{x}_0^{<k})}_{\text{Full-precision (Clean)}} \cdot \underbrace{\hat{p}_\theta(\hat{\boldsymbol{x}}_0^i \mid \boldsymbol{x}_0^{<i})}_{\text{Quantize i-th frame}} \cdot \underbrace{\prod_{k=i+1}^{N} p_\theta(\hat{\boldsymbol{x}}_0^k \mid \boldsymbol{x}_0^{1:i-1}, \  \hat{\boldsymbol{x}}_0^{i:k-1})}_{\text{Full-precision (Noisy)}} \ ,
\end{align}
}

where $\hat{p}_{\theta}$ represents the model in the quantized state, and $\hat{x}$ means this frame is influenced by quantization errors. Here, the quantized model is implemented without reconstruction in ~\Cref{eq:recon}. Note that for $k>i$, the model reverts to full-precision, but the generated frames are still impacted since they are conditioned on the quantized $i$-th and subsequent frames. Finally, the $i$-th frame sensitivity $\alpha_i$ is calculated as the quality degradation caused by quantization, i.e., the distance between the original video $P(\boldsymbol{x}_0^{1:N})$ and the quantized one $\hat{P}_i(\hat{\boldsymbol{x}}_0^{1:N})$, which can be formulated as:

{
\setlength{\abovedisplayskip}{-0.8em}
\setlength{\belowdisplayskip}{0.2em}
\begin{align}
\alpha_i = d(P(\boldsymbol{x}_0^{1:N}), \hat{P}_i(\hat{\boldsymbol{x}}_0^{1:N})). 
\end{align}
}

\begin{figure*}[t]
    \centering
    \includegraphics[width=0.7\linewidth]{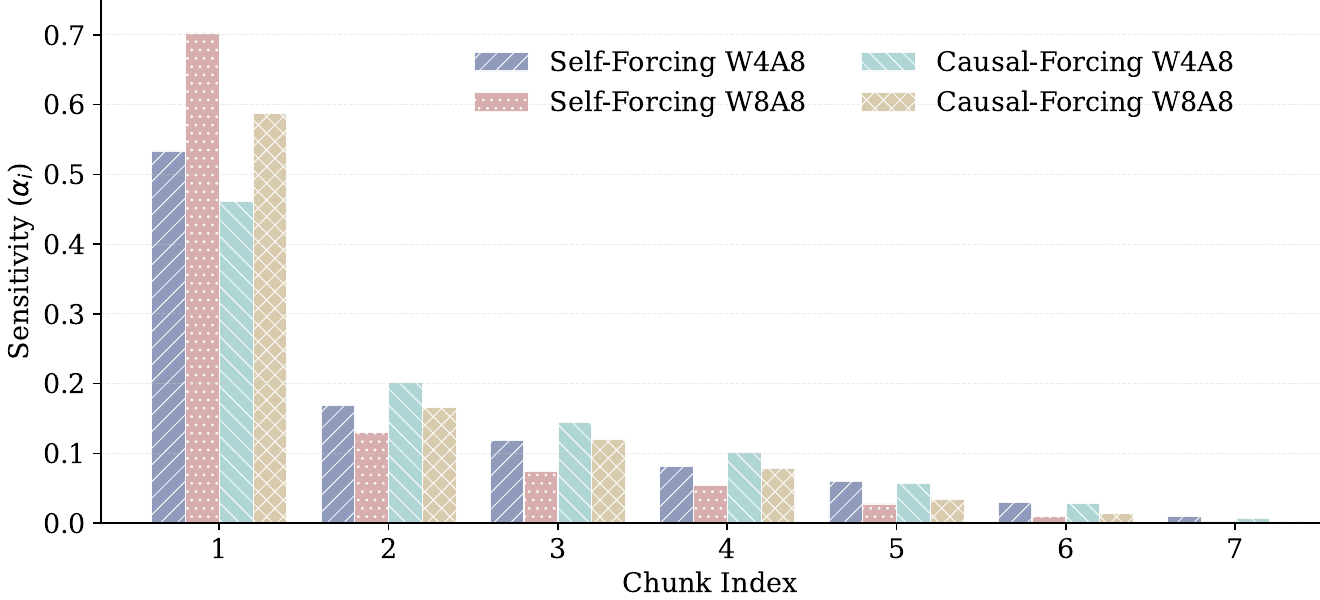}
    \caption{Quantization sensitivity patterns in autoregressive video diffusion models, with scores normalized to sum to 1.}
    \label{fig:sensitivity}
    \vspace{-0.5em}
\end{figure*}

In practice, we compute the distance using the mean-squared error (MSE) in the latent space. We use chunk-wise model following self-forcing~\citep{huang2025selfforcing}, where each chunk contains several frames. Through experiments on 100 videos with different prompts, we obtain the sensitivity pattern shown in~\Cref{fig:sensitivity}. The sensitivity varies significantly across chunks, exhibiting an exponential-like decay. 
For example, the sensitivity score of chunk 1 of self-forcing (W8A8) is 0.70, while the last chunk is less than 0.01. 
The finding indicates that treating all frames equally for quantization calibration is not optimal. Therefore, we use the sensitivity as the loss-weighting coefficients for the quantization reconstruction process. \textbf{The new reconstruction objective is:}

{
\setlength{\abovedisplayskip}{-0.7em}
\setlength{\belowdisplayskip}{0.3em}
\begin{align}
\mathcal{L}_{ours} =
\mathbb{E}_{\mathbf{X} \sim \mathcal{D}_{cal}, \ i \sim \mathcal{U}(1, N)}
\left[
\alpha_i \left\| \mathbf{X}^{(i)} \mathbf{W} - Q(\mathbf{X}^{(i)}) Q(\mathbf{W}) \right\|_\text{F}^2
\right] \ ,
\label{eq:frame_weighting}
\end{align}
}

where $ \mathbf{X}^{(i)}$ means that the activation is obtained from the generation process of the $i$-th frame.

\begin{figure*}[t]
    \centering
    \includegraphics[width=1\linewidth]{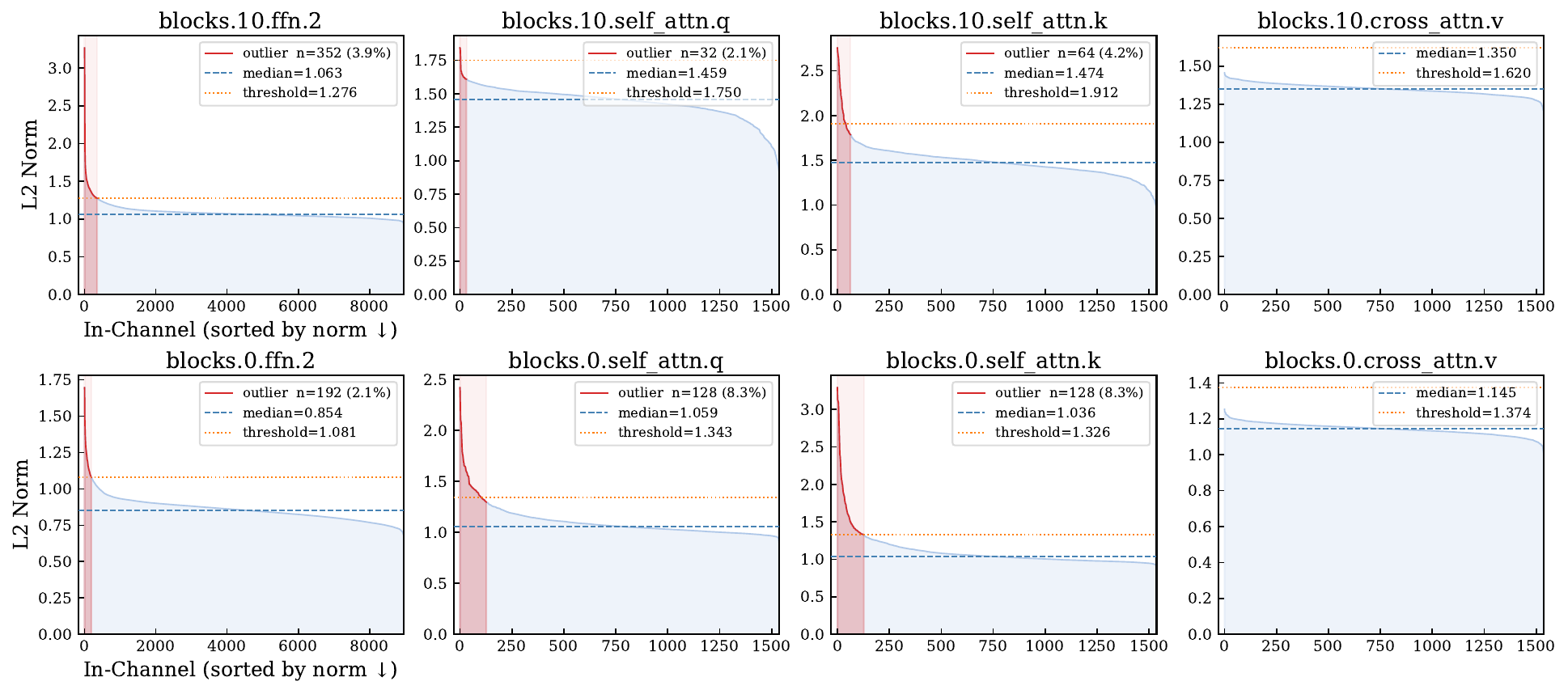}
    \caption{The outlier patterns in autoregressive video diffusion models. The x-axis denotes input channel index sorted in descending L2 norm, and the y-axis denotes the corresponding L2 norm values. Outlier channels identified by our method (\S\ref{sec:dual_scale}, \Cref{eq:new_threshold}) are highlighted in red. In practice, we further align the number of detected outlier channels to a multiple of 32 for hardware-friendly deployment. Additional samples are provided in the Appendix \S\ref{sec:appendix:visual_outliers}. }
    \label{fig:outlier_patterns}
    \vspace{-0.5em}
\end{figure*}

\subsection{Outlier-aware Adaptive Dual-scale Quantization}
\label{sec:dual_scale}

We delve into the weight distributions of autoregressive video diffusion models. Concretely, we collect the statistics of input-channel-wise magnitudes for every layer, as demonstrated in~\Cref{fig:outlier_patterns}.
We compute the per-channel L2 norms and sort them in descending order, from which we can draw the following observations.

\begin{enumerate}
\renewcommand{\labelenumi}{(\roman{enumi})}
    \item \textit{There exist outlier channels which only account for a small fraction but possess obviously larger magnitudes than normal channels.}
    \item \textit{The outlier patterns are highly \textbf{heterogeneous}, varying significantly across different layer types and block depths. For example, the second FFN layers (ffn.2) are prominent, while the cross-attention value projections (cross\_attn.v) are smooth.}
\end{enumerate}

\textbf{Addressing outliers with dual-scale quantization}. Observation (i) reveals that there is ample room to improve quantization quality by addressing these outlier channels. 
Let us start by revisiting why the outliers are not welcome in quantization. The total quantization error $\epsilon$ consists of two components, i.e., the clipping error and the rounding error. The outliers mainly undermine quantization through increasing the rounding error. For example, in symmetric quantization, we have the scaling factor $s=\frac{\max(|x|)}{2^{b-1}-1}$. Let $\hat{x}$ denote the de-quantized value of $x$. From~\Cref{eq:quantization}, we can derive:

{
\setlength{\abovedisplayskip}{-0.8em}
\setlength{\belowdisplayskip}{0.2em}
\begin{align}
\hat{x} = & (x_q-z)\cdot s=[(\operatorname{round}\left(\frac{x}{s}\right)+z)-z] \cdot s = \operatorname{round}\left(\frac{x}{s}\right) \cdot s \ ,  \\
& \mathbb{E}[\epsilon]= \mathbb{E}[|\hat{x}-x|] = \mathbb{E}[ | \operatorname{round}\left(\frac{x}{s}\right)-\frac{x}{s}| \cdot s] = \frac{1}{4}s \ . \label{eq:round_error}
\end{align}
}

Here, we assume the rounding error follows the uniform distribution. Outliers inflate $\max|x|$ and consequently lead to a larger scaling factor $s$. As shown in~\Cref{eq:round_error}, this will lead to a higher quantization error.
To address this problem, we propose a dual-scale quantization strategy to isolate outlier channels from normal channels, thereby preventing them from inflating the quantization errors, which can be formulated as:

{
\vspace{-1.5em}
\begin{align}
Q_{\text{dual}}(\mathbf{W})=\Big[ \, Q_{\text{outlier}}(\mathbf{W}_{\text{outliers}}) \; | \;
Q_{\text{normal}}(\mathbf{W}_{\text{normal}}) \Big] \ ,
\label{eq:dual_scale}
\end{align}
\vspace{-1.0em}
}

where $[\,\cdot | \cdot\,]$ denotes concatenation along the input-channel dimension, and $Q_{\text{outlier}}$ and $Q_{\text{normal}}$ are two independent quantizers, for outlier and normal channels respectively. The separate quantizer results in a lower scaling factor for normal channels, and theoretically reduce quantization errors according to~\Cref{eq:round_error}. We also discuss and compare related outlier-handling approaches in Appendix \S\ref{sec:appendix_discuss}.

\textbf{Adaptively detecting heterogeneous outlier patterns.} However, Observation (ii) indicates that the outlier patterns are heterogeneous across layers. Some layers (e.g., ffn.2) manifest significant outliers, while certain layers (e.g., cross\_attn.v) exhibit smooth distributions. 
This disparity raises a critical question: How to determine whether there exists an outlier pattern and how many top channels should be regarded as outliers? Manually tuning is labor-intensive and lacks generalizability. To achieve automatic and adaptive outlier detection, we employ the Modified Z-score~\citep{iglewicz1993volume}. Given the L2 norm vector $\boldsymbol{v} \in \mathbb{R}^{d_{\text{in}}}$, we first compute the  Median Absolute Deviation (MAD):

{
\setlength{\abovedisplayskip}{-0.8em}
\setlength{\belowdisplayskip}{-0em}
\begin{align}
\text{MAD} = \text{median}\left(|v_i - \tilde{v} | \right), \text{where} \ \ \tilde{v} = \text{median}(\boldsymbol{v}).
\label{eq:mad}
\end{align}
}

The Modified Z-score for each channel is formulated as:

{
\setlength{\abovedisplayskip}{-1.0em}
\setlength{\belowdisplayskip}{-0.2em}
\begin{align}
\text{M}_i = 0.6745 \cdot \frac{v_i - \tilde{v}}{\text{MAD}},
\label{eq:modified_zscore}
\end{align}
}

The Modified Z-score measures how far a channel deviates from the median in a normalized manner. Following the standard Modified Z-score criterion, a channel is marked as an outlier when $\text{M}_i$ exceeds a threshold $\tau$, i.e., $0.6745 \cdot \frac{v_i - \tilde{v}}{\text{MAD}}>\tau$, which can be rewritten as:

{
\setlength{\abovedisplayskip}{-0.8em}
\setlength{\belowdisplayskip}{0.2em}
\begin{align}
v_i > \tilde{v} + \frac{\tau}{0.6745} \cdot \text{MAD} \ .
\label{eq:ori_threshold}
\end{align}
}

However, we observe that for certain smooth layers, the MAD can be extremely small, resulting in a low right-hand side of~\Cref{eq:ori_threshold}. This will mark a lot of normal values as outliers, which we refer to as ``false outliers''. To avoid this issue, we introduce a minimum magnitude constraint. \textbf{Finally, a channel is classified as outlier when it satisfies both the Modified Z-score and the minimum magnitude conditions:}

\begin{figure*}[t]
    \centering
    \includegraphics[width=0.9\linewidth]{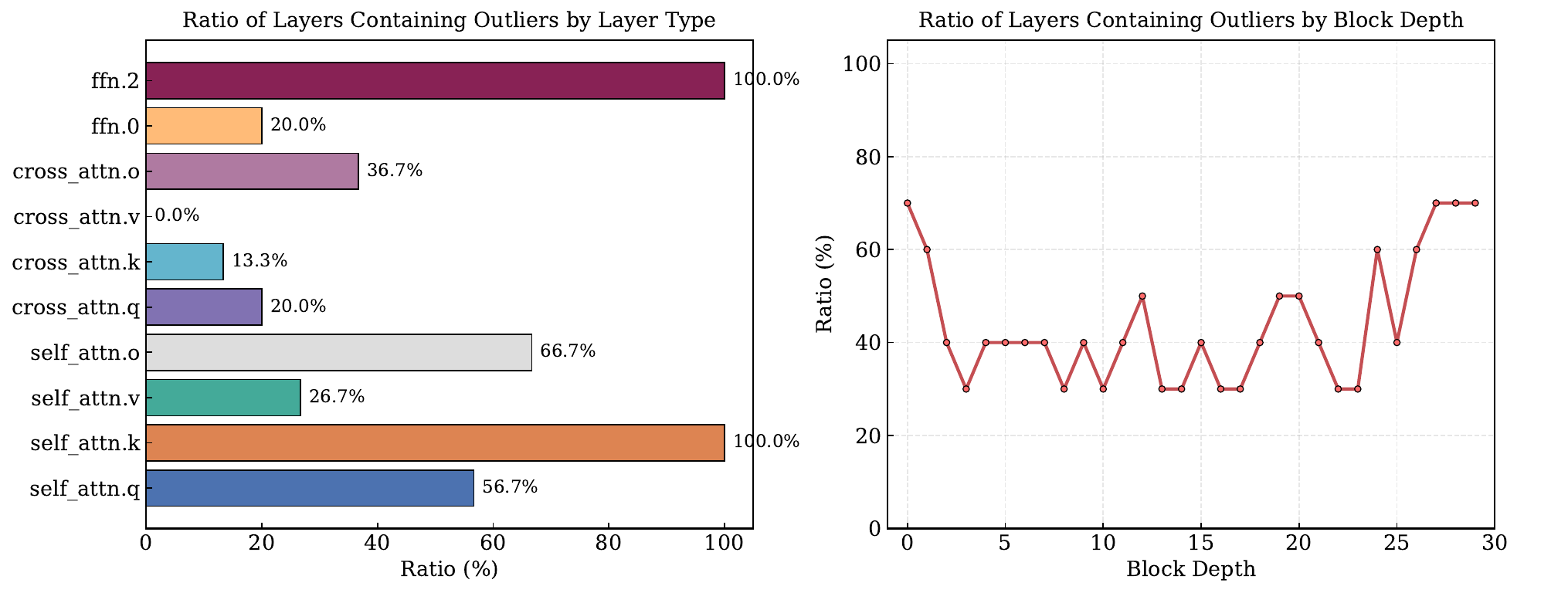}
    \caption{The ratio of layers containing outliers in terms of layer type and block depth. A layer is identified as outlier-containing upon detection of at least one outlier channel.}
    \label{fig:outlier_summary}
    \vspace{-1em}
\end{figure*}

{
\setlength{\abovedisplayskip}{-1.5em}
\setlength{\belowdisplayskip}{0.2em}
\begin{align}
v_i > \max(\tilde{v} + \frac{\tau}{0.6745} \cdot \text{MAD}, \ \ \alpha\cdot\tilde{v})\ ,
\label{eq:new_threshold}
\end{align}
}

where $\tau=3.5$ is the standard Modified Z-score threshold, and $\alpha=1.2$ (default) denotes the minimum ratio relative to the median norm. A layer is considered to contain outlier channels if at least one outlier channel is detected, in which case dual-scale quantization will be applied. \Cref{fig:outlier_summary} shows the proportion of layers containing outliers in terms of layer type and block depth.

\section{Experiments}

\subsection{Experimental Setups}
\label{sec:exp_settings}

\textbf{Models and Baselines.} We use two state-of-the-art autoregressive video diffusion models, i.e., self-forcing~\citep{huang2025selfforcing} and causal-forcing~\citep{zhu2026causal}, and follow their official configurations. Our baselines include five representative quantization paradigms. Specifically, MinMax~\citep{nagel2021white} serves as a vanilla quantization approach, while AdaRound~\citep{nagel2020up} represents a classical reconstruction-based method. 
SmoothQuant~\citep{xiao2023smoothquant} is a widely adopted method for handling activation outliers in transformers via channel-wise scaling. PTQ4DiT~\citep{wu2024ptq4dit} is a framework tailored for diffusion transformers. SVDQuant~\citep{lisvdquant} pioneers in mitigating weight outliers by introducing a low-rank full-precision branch.

\textbf{Quantization Implementation.} In all baselines and our method, we use per-channel quantization for weights and per-tensor static quantization for activations. In the initialization of scaling factors, we search for the optimal percentile of clipping from [0.999, 0.9999, 0.99999]. We choose the extended MovieGenVideoBench prompts~\citep{polyak2024movie,huang2025selfforcing} as calibration data. 
Following previous works~\citep{wu2024ptq4dit,li2023q}, we train adaptive rounding and scaling factors by reconstruction.

\textbf{Benchmark and Metrics.} Following common practice in evaluating video generative models, we evaluate quantized models on the VBench benchmark~\citep{huang2023vbench}. We adopt two types of metrics, i.e., reference-based metrics and reference-free metrics.
\textbf{Reference-based} metrics measure the distance
between videos generated by quantized models and those from the full-precision (FP) model~\citep{zhaovidit,tang2024post}. Specifically, we adopt two popular distance metrics, i.e., FVD~\citep{unterthiner2018towards} and LPIPS~\citep{zhang2018unreasonable}, denoted as FVD-FP~\citep{zhaovidit} and LPIPS-FP in our quantization task, respectively. For reference-free evaluation, we report five VBench quality scores. To ensure reliable evaluation, especially for FVD-FP, we generate videos using all 946 extended VBench prompts.
Empirically, we observe that VBench scores exhibit limited discriminative power for evaluating quantization performance, whereas reference-based metrics are far more sensitive and better aligned with actual quality. Therefore, we primarily rely on \textbf{FVD-FP and LPIPS-FP} for quantitative comparison, while using VBench scores as auxiliary evidence.

\subsection{Main Results}

\definecolor{baselinegray}{gray}{0.93}
\definecolor{oursorange}{RGB}{232,240,248}
\definecolor{metrichead}{RGB}{255,245,235}

{
\begin{table*}[t]
\caption{Quantitative results of causal-forcing~\citep{zhu2026causal}. We use the reference-based metrics, i.e., FVD-FP and LPIPS-FP, as the primary criteria for quantitative comparison, while treating VBench scores as complementary evidence.}
\label{table:main_results_causal_forcing}
\centering
\setlength{\tabcolsep}{4pt}
\resizebox{1.0\textwidth}{!}{%
\begin{tabular}{>{\centering\arraybackslash}c >{\centering\arraybackslash}c cccccc|cc}
\toprule
\multirow{3.5}{*}{Method} & \multirow{3.5}{*}{Bitwidth} & \multicolumn{6}{c}{Reference-free Metrics} & \multicolumn{2}{c}{\cellcolor{metrichead}\textbf{Reference-based Metrics}} \\
\cmidrule(lr){3-8}\cmidrule(l){9-10}
& &
\begin{tabular}[c]{@{}c@{}}Subj.\\ Cons.\end{tabular} &
\begin{tabular}[c]{@{}c@{}}Back.\\ Cons.\end{tabular} &
\begin{tabular}[c]{@{}c@{}}Motion\\ Smooth.\end{tabular} &
\begin{tabular}[c]{@{}c@{}}Aesth.\\ Qual.\end{tabular} &
\begin{tabular}[c]{@{}c@{}}Imag.\\ Qual.\end{tabular} &
\begin{tabular}[c]{@{}c@{}}Avg. \end{tabular} &
\cellcolor{metrichead}\textbf{FVD-FP $\downarrow$} &
\cellcolor{metrichead}\textbf{LPIPS-FP $\downarrow$} \\
\midrule

\rowcolor{baselinegray}
Bfloat16      & 16/16 & 96.91 & 96.59 & 98.47 & 62.88 & 71.80 & 85.33 & 0.00 & 0.00 \\ \midrule

MinMax        &  & 95.36 & 95.37 & 98.82 & 59.08 & 68.41 & 83.41 & 279.61 & 0.505 \\
Adaround      &  & 96.38 & 96.02 & 98.56 & 61.71 & 71.40 & 84.81 & 143.49 & 0.463 \\
SmoothQuant   &  & 96.56 & 95.84 & 98.71 & 60.77 & 71.25 & 84.63 & 220.94 & 0.507 \\
PTQ4DiT       &  & 96.71 & 96.07 & 98.70 & 61.65 & 71.69 & 84.96 & 141.03 & 0.470 \\
SVDQuant      &  & 96.24 & 95.94 & 98.81 & 59.98 & 70.35 & 84.26 & 135.62 & 0.492 \\
\rowcolor{oursorange}
Q-ARVD (Ours) & \multirow{-6}{*}{W4A8} & 96.74 & 96.33 & 98.56 & 61.92 & 71.23 & 84.96 & \textbf{106.04} & \textbf{0.452} \\ \midrule

MinMax        &  & 96.71 & 96.45 & 98.46 & 62.33 & 72.16 & 85.22 & 67.65  & 0.358 \\
Adaround      &  & 96.95 & 96.60 & 98.52 & 62.68 & 71.93 & 85.34 & 62.57  & 0.341 \\
SmoothQuant   &  & 96.81 & 96.50 & 98.64 & 62.40 & 72.10 & 85.29 & 69.67  & 0.360 \\
PTQ4DiT       &  & 96.97 & 96.66 & 98.52 & 62.60 & 72.08 & 85.37 & 63.21  & 0.341 \\
SVDQuant      &  & 96.80 & 96.55 & 98.43 & 62.66 & 72.07 & 85.30 & 63.52  & 0.361 \\
\rowcolor{oursorange}
Q-ARVD (Ours) & \multirow{-6}{*}{W8A8} & 96.98 & 96.59 & 98.52 & 62.66 & 72.04 & 85.36 & \textbf{61.67} & \textbf{0.335} \\ \midrule

MinMax        &  & 94.82 & 94.95 & 98.73 & 58.27 & 65.77 & 82.51 & 375.82 & 0.544 \\
Adaround      &  & 96.40 & 95.96 & 98.58 & 61.62 & 69.96 & 84.50 & 233.43 & 0.507 \\
SmoothQuant   &  & 96.57 & 95.61 & 98.75 & 59.22 & 69.72 & 83.97 & 326.97 & 0.539 \\
PTQ4DiT       &  & 96.21 & 95.28 & 98.55 & 60.53 & 70.75 & 84.26 & 244.95 & 0.527 \\
SVDQuant      &  & 96.13 & 95.76 & 98.65 & 59.20 & 68.76 & 83.70 & 210.28 & 0.532 \\
\rowcolor{oursorange}
Q-ARVD (Ours) & \multirow{-6}{*}{W4A6} & 97.32 & 96.66 & 98.82 & 62.55 & 70.86 & 85.24 & \textbf{140.38} & \textbf{0.486} \\

\bottomrule
\end{tabular}
}
\vspace{-0.5em}
\end{table*}
}

\begin{figure*}[t]
     \vspace{-1em}
    \centering
    \includegraphics[width=0.95\linewidth]{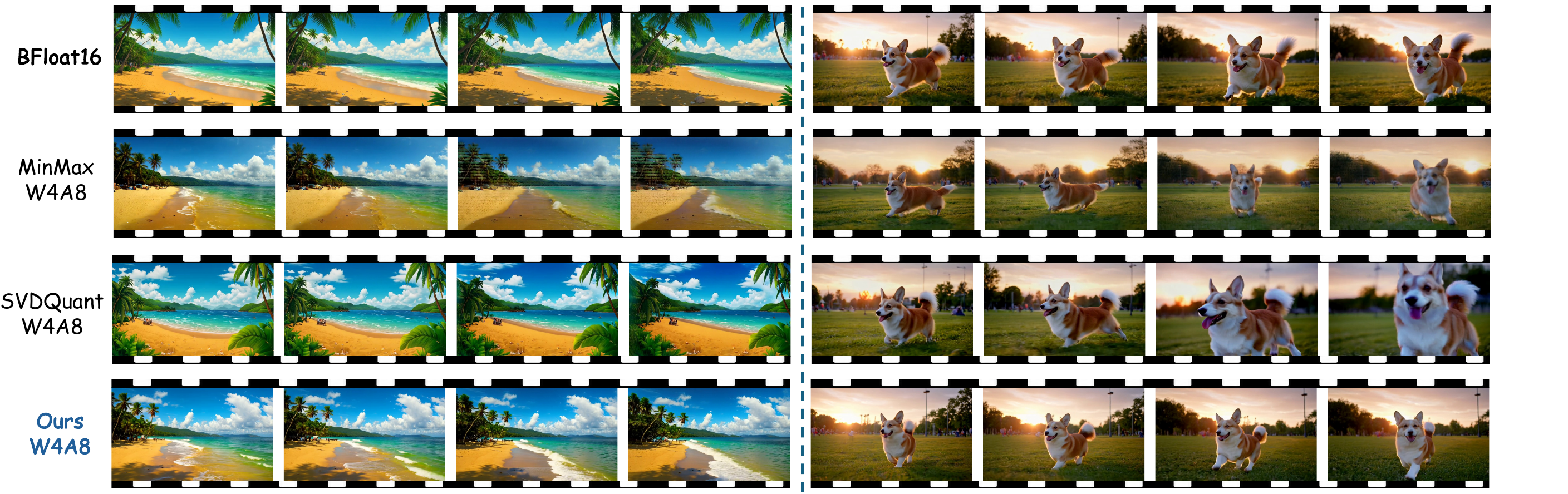}
    \caption{The visual comparison of the self-forcing model with
    W4A8. Additional samples are presented in the Appendix (\S\ref{sec:appendix:visual_samples} \Cref{fig:appendix_visual_self_w4a8}- \Cref{fig:appendix_visual_causal_w8a8}).}
    \label{fig:visual_compare_main}
    \vspace{0em}
\end{figure*}

\textbf{Main comparison and visual results.} \Cref{table:main_results_causal_forcing} and \Cref{table:main_results_self_forcing} show the quantitative results on causal-forcing and self-forcing, respectively.
We test three different bitwidths, i.e., W8A8, W4A8, and W4A6, with increasing quantization difficulty. The results show that Q-ARVD consistently achieves the best FVD-FP and LPIPS-FP scores, outperforming all baselines by a clear margin. The improvement is more pronounced under low-bit settings (i.e., W4A8 and W4A6), where the outlier issue becomes more severe.
Moreover, \Cref{fig:visual_compare_main} shows the visual results. MinMax suffers from significant accumulated errors, leading to severe degradation in frame quality over time. SVDQuant introduces noticeable semantic changes compared to the original Bfloat16 video, such as the style and viewpoint of the beach, and the appearance and posture of the dog. In contrast, our method maintains high video quality consistently across the full temporal span. More visual examples can be found in the Appendix (\S\ref{sec:appendix:visual_samples} \Cref{fig:appendix_visual_self_w4a8}- \Cref{fig:appendix_visual_causal_w8a8}). We also validate the practical deployment of the W8A8 model using Triton~\citep{tillet2019triton}, and observe a 1.30x latency speedup on the NVIDIA A6000 GPU and a 1.97x model size reduction, with a batch size of 2 and default configurations.

{
\begin{table*}[t]

\caption{Quantitative results of self-forcing~\citep{huang2025selfforcing}.}
\label{table:main_results_self_forcing}
\centering
\setlength{\tabcolsep}{5pt}
\resizebox{1.0\textwidth}{!}{%
\begin{tabular}{>{\centering\arraybackslash}c >{\centering\arraybackslash}c cccccc|cc}
\toprule
Method & Bitwidth &
\begin{tabular}[c]{@{}c@{}}Subj.\\ Cons.\end{tabular} &
\begin{tabular}[c]{@{}c@{}}Back.\\ Cons.\end{tabular} &
\begin{tabular}[c]{@{}c@{}}Motion\\ Smooth.\end{tabular} &
\begin{tabular}[c]{@{}c@{}}Aesth.\\ Qual.\end{tabular} &
\begin{tabular}[c]{@{}c@{}}Imag.\\ Qual.\end{tabular} &
\begin{tabular}[c]{@{}c@{}}Avg.\end{tabular} &
\cellcolor{metrichead}\textbf{FVD-FP $\downarrow$} &
\cellcolor{metrichead}\textbf{LPIPS-FP $\downarrow$} \\
\midrule

\rowcolor{baselinegray}

Bfloat16      & 16/16 & 97.25 & 96.68 & 98.99 & 62.49 & 71.35 & 85.35 & 0.00 & 0.00 \\ \midrule
MinMax        &  & 95.82 & 95.00 & 98.92 & 58.39 & 68.53 & 83.33 & 260.14 & 0.514 \\
Adaround      &  & 96.89 & 96.05 & 98.96 & 60.79 & 70.12 & 84.56 & 156.70    & 0.474 \\
SmoothQuant   &  & 96.72 & 95.82 & 98.89 & 59.24 & 71.05 & 84.34 & 220.30    & 0.513 \\
PTQ4DiT       &  & 97.17 & 96.20 & 98.96 & 61.32 & 71.11 & 84.95 & 124.20 & 0.477 \\
SVDQuant      &  & 96.41 & 95.56 & 98.78 & 58.62 & 70.65 & 84.00 & 150.03 & 0.502 \\
\rowcolor{oursorange}
Q-ARVD (Ours) & \multirow{-6}{*}{W4A8} & 97.05 & 96.27 & 98.95 & 61.00 & 70.71 & 84.80 & \textbf{116.26} & \textbf{0.466} \\ \midrule

MinMax        &  & 97.31 & 96.64 & 98.99 & 61.83 & 71.80 & 85.31 & 77.65  & 0.351 \\
Adaround      &  & 97.29 & 96.62 & 99.00 & 62.03 & 71.44 & 85.28 & 68.24  & 0.334 \\
SmoothQuant   &  & 97.31 & 96.65 & 99.00 & 61.96 & 71.75 & 85.33 & 81.08  & 0.354 \\
PTQ4DiT       &  & 97.33 & 96.69 & 99.01 & 62.07 & 71.43 & 85.31 & 68.47  & 0.333 \\
SVDQuant      &  & 97.29 & 96.67 & 99.01 & 61.97 & 71.72 & 85.33 & 74.87  & 0.349 \\
\rowcolor{oursorange}
Q-ARVD (Ours) & \multirow{-6}{*}{W8A8} & 97.26 & 96.62 & 98.99 & 62.13 & 71.49 & 85.30 & \textbf{64.51} & \textbf{0.327} \\ \midrule

MinMax        &  & 96.11 & 94.95 & 99.04 & 57.84 & 67.30 & 83.05 & 321.32 & 0.534 \\
Adaround      &  & 96.68 & 95.59 & 99.06 & 59.79 & 67.85 & 83.79 & 224.95 & 0.515 \\
SmoothQuant   &  & 96.96 & 95.77 & 98.97 & 58.61 & 70.06 & 84.07 & 284.80 & 0.535 \\
PTQ4DiT       &  & 97.16 & 95.95 & 98.98 & 60.66 & 70.17 & 84.58 & 174.17 & 0.515 \\
SVDQuant      &  & 96.67 & 95.71 & 98.70 & 58.29 & 70.09 & 83.89 & 215.32 & 0.542 \\
\rowcolor{oursorange}
Q-ARVD (Ours) & \multirow{-6}{*}{W4A6} & 97.39 & 96.30 & 99.03 & 60.79 & 70.33 & 84.77 & \textbf{146.01} & \textbf{0.498} \\

\bottomrule
\end{tabular}
}
\end{table*}
}

\begin{table*}[t]
\centering
\vspace{-1.1em}
\begin{minipage}[t]{0.535\textwidth}
    \centering
    \caption{Ablation study of frame weighting and dual scale quantization using self-forcing.}
    \label{tab:ablation-1}
    \resizebox{\linewidth}{!}{%
    \begin{tabular}{c c cc cc}
      \toprule
      \multirow{2}{*}{Dual Scale} & \multirow{2}{*}{Frame Weighting} 
      & \multicolumn{2}{c}{W4A8} 
      & \multicolumn{2}{c}{W8A8} \\
      \cmidrule(lr){3-4} \cmidrule(lr){5-6}
       & & FVD$\downarrow$ & LPIPS$\downarrow$ & FVD$\downarrow$ & LPIPS$\downarrow$ \\
      \midrule
      \xmark & \xmark & 156.70 & 0.474 & 68.24 & 0.334 \\
      \xmark & \cmark & 147.16 & \textbf{0.465} & 65.39 & \textbf{0.325} \\
      \cmark & \xmark & 121.83 & 0.469 & 67.48 & 0.332 \\
      \cmark & \cmark & \textbf{116.26} & 0.466 & \textbf{64.51} & 0.327 \\
      \bottomrule
    \end{tabular}%
    }
\end{minipage}
\hfill
\begin{minipage}[t]{0.40\textwidth}
    \centering
    \caption{Comparison of different frame-weighting strategies (self-forcing).}
    \label{tab:weighting_ablation}
    \resizebox{\linewidth}{!}{%
    \begin{tabular}{ccc}
      \toprule
      Strategy & FVD $\downarrow$ & LPIPS $\downarrow$ \\
      \midrule
      Uniform (no weighting) & 121.83 & 0.469 \\
      Heuristic Exp $2^{-i}$  & 119.61 & 0.471 \\
      Reverse & 123.72 & 0.476 \\
      \midrule
      Ours & \textbf{116.26} & \textbf{0.466} \\
      \bottomrule
    \end{tabular}
    }
\end{minipage}
\vspace{-1.5em}
\end{table*}

\textbf{Discussion of VBench scores in quantization}. Our experiments reveal that standard VBench metrics have limited discriminability and can even exhibit counterintuitive behavior when evaluating quantized video models. For example, in~\Cref{table:main_results_self_forcing}, the motion smoothness score varies only marginally (98.70--99.06), and the average VBench scores of all W8A8 models fall within a narrow range of 85.28--85.33, while FVD-FP and LPIPS-LP changes substantially across settings.
More surprisingly, our lower-precision W4A6 Q-ARVD outperforms W8A8, W4A8, and even BF16 in metrics such as subject consistency and motion smoothness.
A similar phenomenon is observed in~\Cref{table:main_results_causal_forcing}, for instance, the average score of W4A6 Q-ARVD surpasses W4A8.
To systematically assess metric reliability, we introduce a unified \textit{Discriminability Score} (DS), which measures both \emph{sensitivity} and \emph{faithfulness}. Sensitivity is quantified by the coefficient of variation (CV)~\citep{abdi2010coefficient}, $\mathrm{CV}_m = \sigma_m / \mu_m$, where higher values indicate stronger responsiveness to quality differences. Faithfulness is measured by the proposed bitwidth-order agreement (BOA), which evaluates whether a metric follows the expected quantization ordering, i.e., BF16 $\succ$ W8A8 $\succ$ W4A8 $\succ$ W4A6. 
Formally, $\mathrm{BOA}_m = \frac{1}{N} \sum_{i=1}^{N} \mathbb{I} \big( m^{\text{BF16}} \succ m_i^{\text{W8A8}} \succ m_i^{\text{W4A8}} \succ m_i^{\text{W4A6}} \big)$, where $N$ is the number of quantization methods ($N=6$ in our setting). 
A higher BOA indicates better consistency with the bitwidth ordering.
We define the final score as $\mathrm{DS}_m = \mathrm{CV}_m \cdot \mathrm{BOA}_m$. As shown in \Cref{fig:metric_analysis}, FVD-FP and LPIPS-FP achieve substantially higher CV, BOA, and DS scores, indicating better sensitivity and consistency with true model quality. In contrast, VBench metrics show low dispersion and poor alignment with the expected bitwidth ordering, suggesting that commonly used metrics may be unreliable for quantization evaluation without careful validation.

\subsection{Ablation Study}

\textbf{Effectiveness of frame weighting and dual scale quantization.} We validate the effectiveness of the final-quality guided frame-weighting (Frame Weighting) and outlier-aware adaptive dual-scale quantization (Dual Scale), as presented in~\Cref{tab:ablation-1}. Both frame weighting and dual-scale quantization individually bring performance gains. Specifically, frame weighting delivers more pronounced improvements under the high-precision W8A8 setting. In contrast, dual-scale quantization exhibits superior efficacy in the low bit-width W4A8 setting. This is because low-precision weight representation is more vulnerable to performance degradation caused by outliers, making the use of outlier-aware dual-scale quantization particularly critical.
Finally, the joint integration of the two modules achieves the best overall performance.

\begin{figure}[t]
    \centering
    \begin{minipage}[t]{0.47\linewidth}
        \centering
        \includegraphics[width=\linewidth]{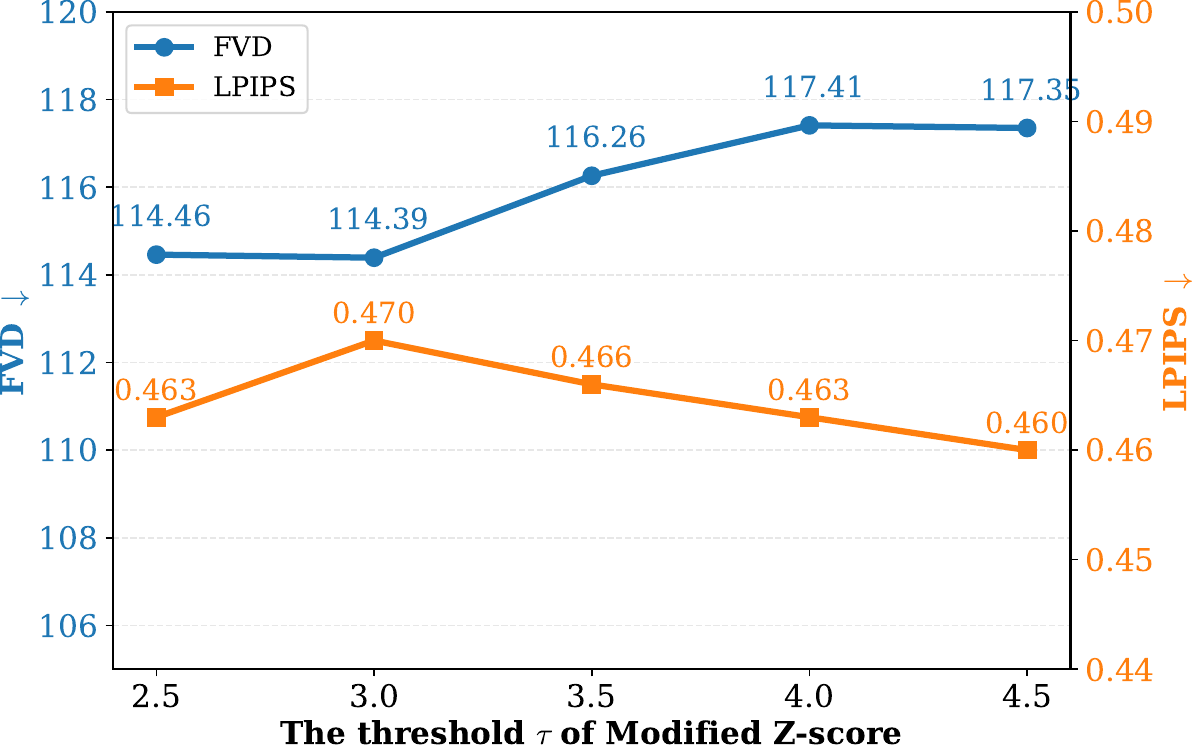}
        \caption{The sensitivity to threshold $\tau$ of Modified Z-score using self-forcing.}
        \label{fig:ablation-2}
    \end{minipage}
    \hspace{1em} 
    \begin{minipage}[t]{0.45\linewidth}
        \centering
        \includegraphics[width=\linewidth]{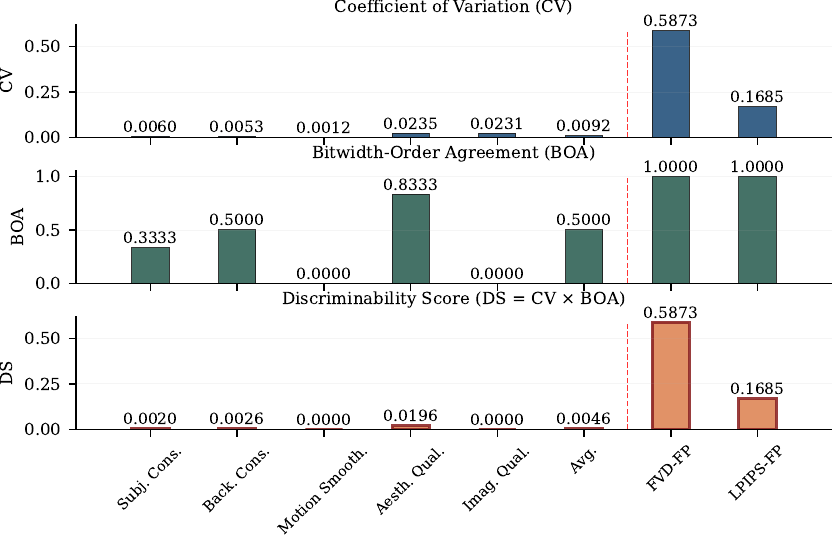}
        \caption{The coefficient of variation, bitwidth-order agreement and discriminability score of each metric. Use data of~\Cref{table:main_results_causal_forcing}.}
        \label{fig:metric_analysis}
    \end{minipage}
    \vspace{-2em}
\end{figure}

\textbf{The sensitivity to threshold $\tau$ of Modified Z-score.} We further conduct sensitivity experiments to validate the robustness of the outlier-aware adaptive dual-scale module, focusing on the threshold $\tau$ of the Modified Z-score. As illustrated in~\Cref{fig:ablation-2}, we vary $\tau$ from 2.5 to 4.5. The results show the FVD-FP score fluctuates slightly within a narrow range of 114.39 to 117.41, and the LPIPS-FP score remains stable between 0.460 and 0.470 across all threshold settings. The slight performance variations demonstrate that our dual-scale module is robust to the selection of $\tau$.

\textbf{Compared to heuristic frame-weighting strategies.} 
We compare our final-quality guided weighting with a heuristic exponential decay of $2^{-i}$ and the reversed version of our weighting. As shown in~\Cref{tab:weighting_ablation}, the heuristic decay is inferior to our final-quality guided strategy, and the reversed version even underperforms the uniform baseline, as it incorrectly emphasizes later frames.

\section{Conclusion}

We propose Q-ARVD, the first quantization framework tailored for autoregressive video diffusion models. Q-ARVD introduces a final-quality guided frame-weighting mechanism to handle the unbalanced frame-wise quantization sensitivity, and an outlier-aware adaptive dual-scale strategy to address the heterogeneous outlier patterns. Extensive quantitative and qualitative experiments show the superiority and rationality of our designs. 
We hope our method can shed light on frame-wise sensitivity-aware quantization and a new path for detecting and handling outliers.

\clearpage
{
    \bibliographystyle{acl_natbib}
    \bibliography{reference}

}

\newpage
\appendix

\section{Additional Visualization of Outlier Patterns}
\label{sec:appendix:visual_outliers}

\begin{figure*}[htbp]
    \centering
    \includegraphics[width=0.99\linewidth]{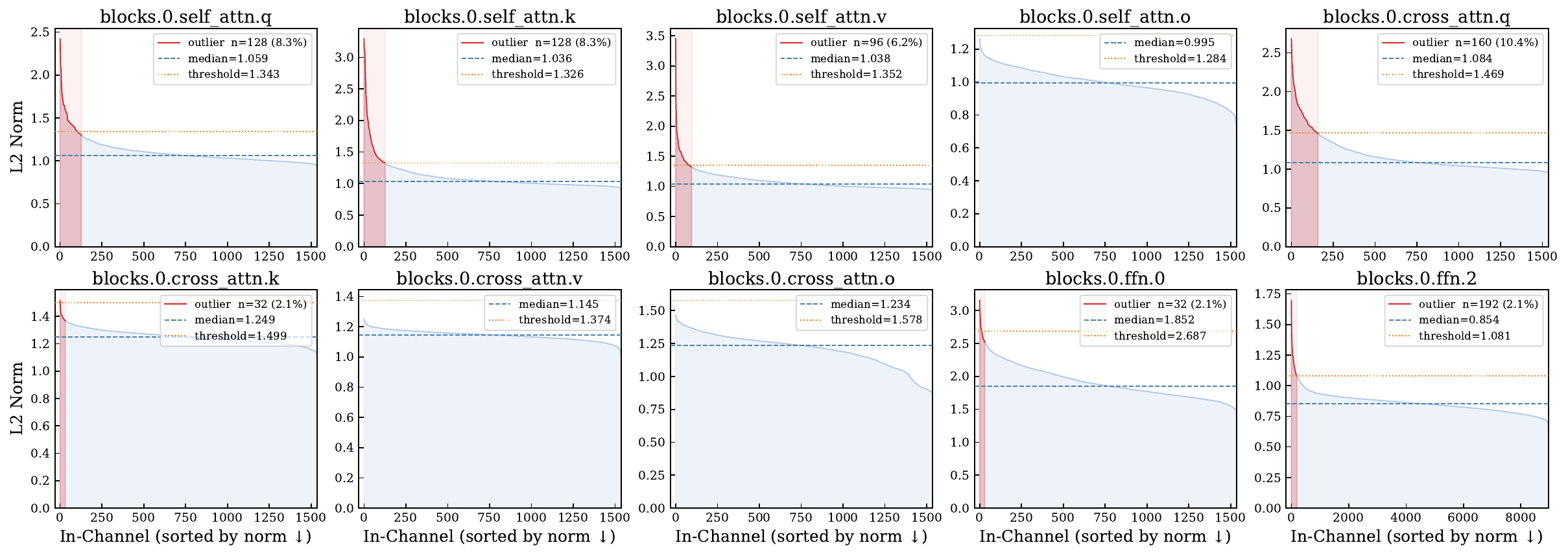}
    \caption{Outlier patterns of all layers in block 0.}
    \label{fig:appendix_outliers_block0}
\end{figure*}

\begin{figure*}[htbp]
    \centering
    \includegraphics[width=0.99\linewidth]{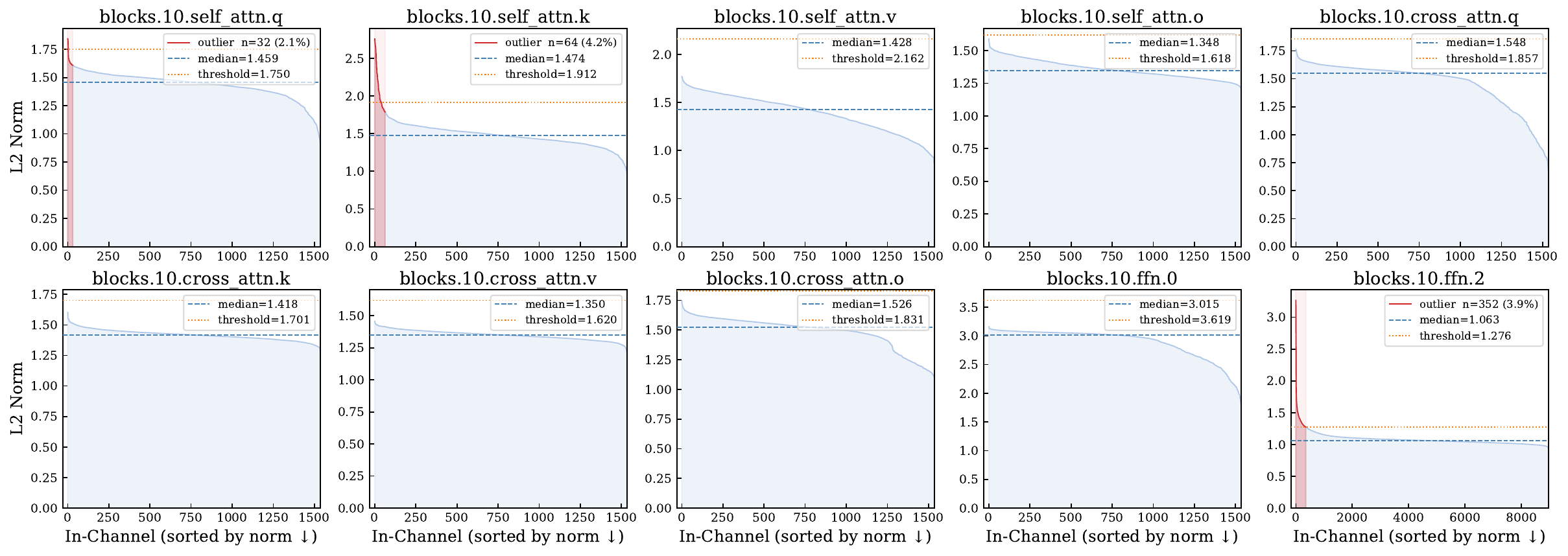}
    \caption{Outlier patterns of all layers in block 10.}
    \label{fig:appendix_outliers_block10}
\end{figure*}

\begin{figure*}[htbp]
    \centering
    \includegraphics[width=0.99\linewidth]{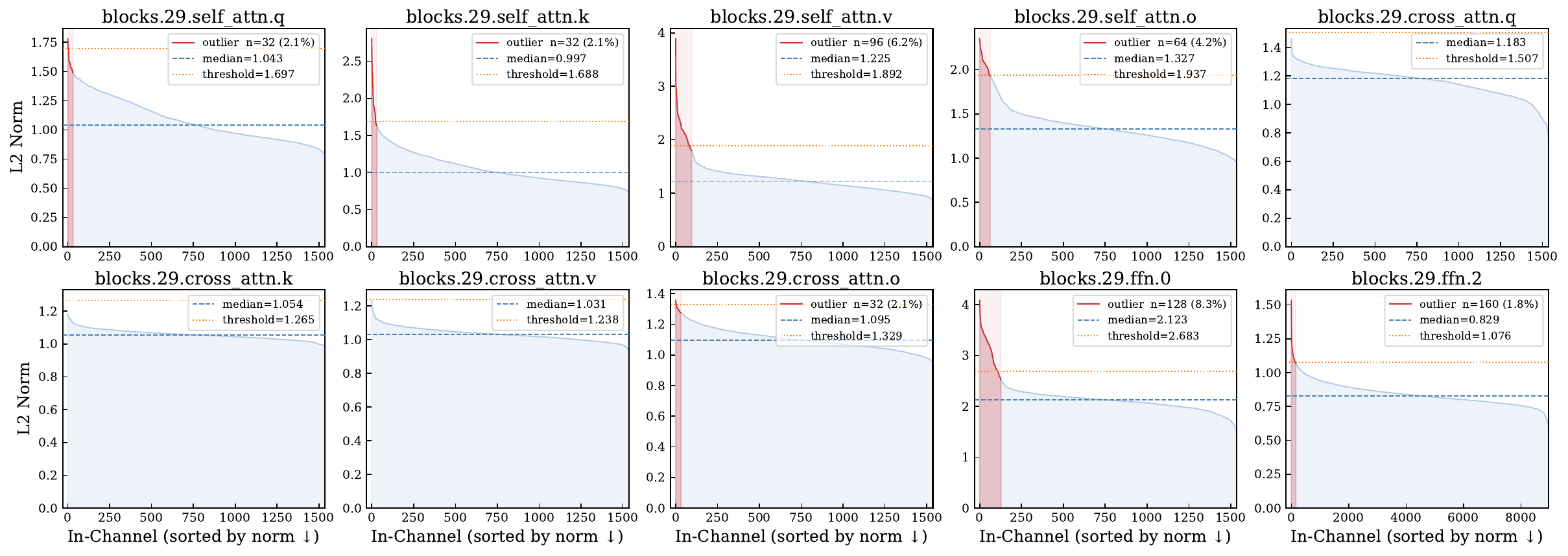}
    \caption{Outlier patterns of all layers in block 29.}
    \label{fig:appendix_outliers_block29}
\end{figure*}

\Cref{fig:appendix_outliers_block0}, \Cref{fig:appendix_outliers_block10}, and \Cref{fig:appendix_outliers_block29} show the outlier patterns of all 10 layers in block 0, 10, and 29.

\section{Additional Samples for Visual Comparison}
\label{sec:appendix:visual_samples}

\begin{figure*}[htbp]
    \centering
    \includegraphics[width=\linewidth]{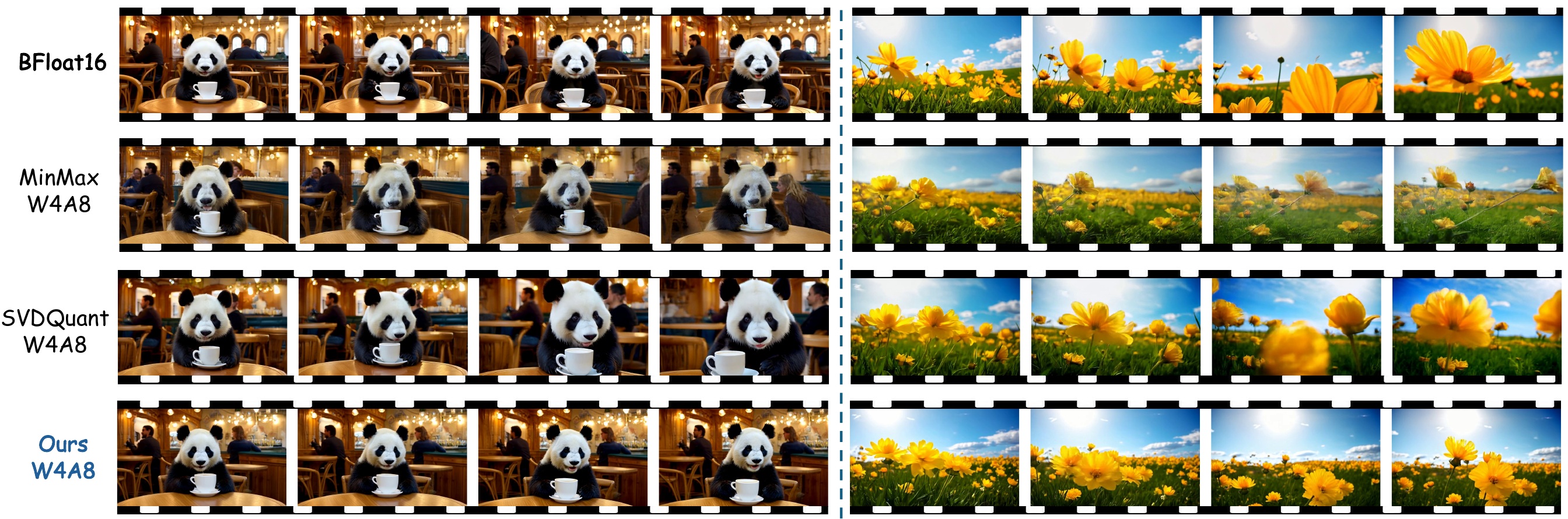}
    \caption{Visual comparison of the self-forcing model using W4A8.}
    \label{fig:appendix_visual_self_w4a8}
\end{figure*}

\begin{figure*}[htbp]
    \centering
    \includegraphics[width=\linewidth]{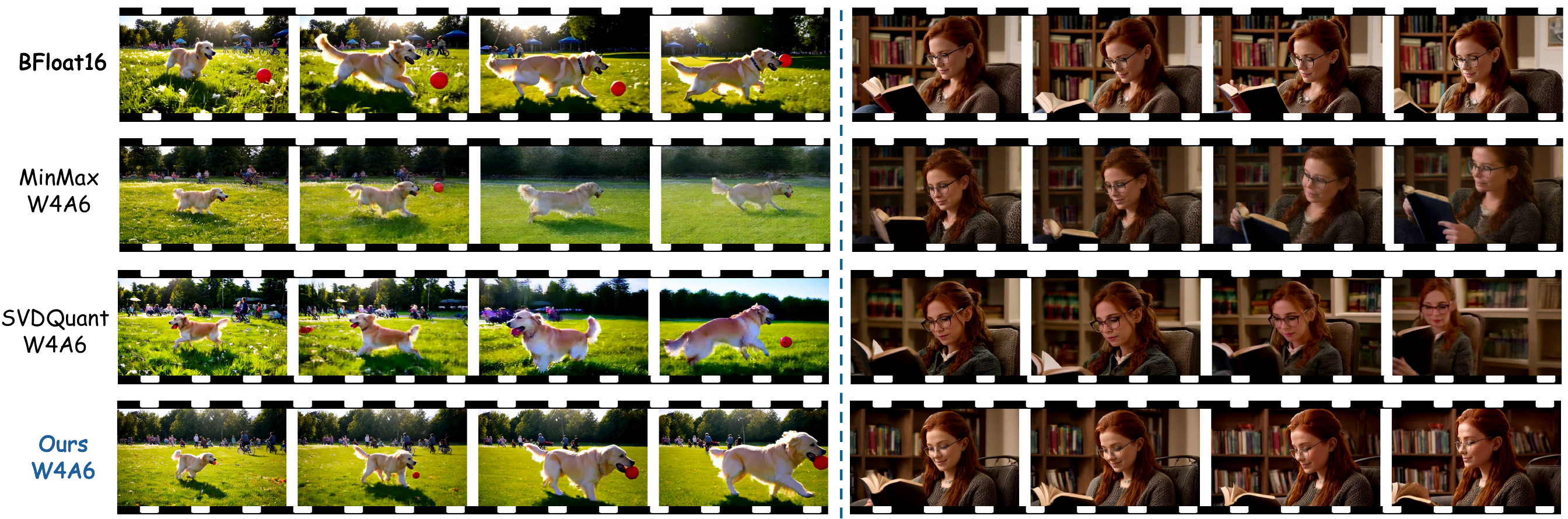}
    \caption{Visual comparison of the self-forcing model using W4A6.}
    \label{fig:appendix_visual_self_w4a6}
\end{figure*}

\begin{figure*}[htbp]
    \centering
    \includegraphics[width=\linewidth]{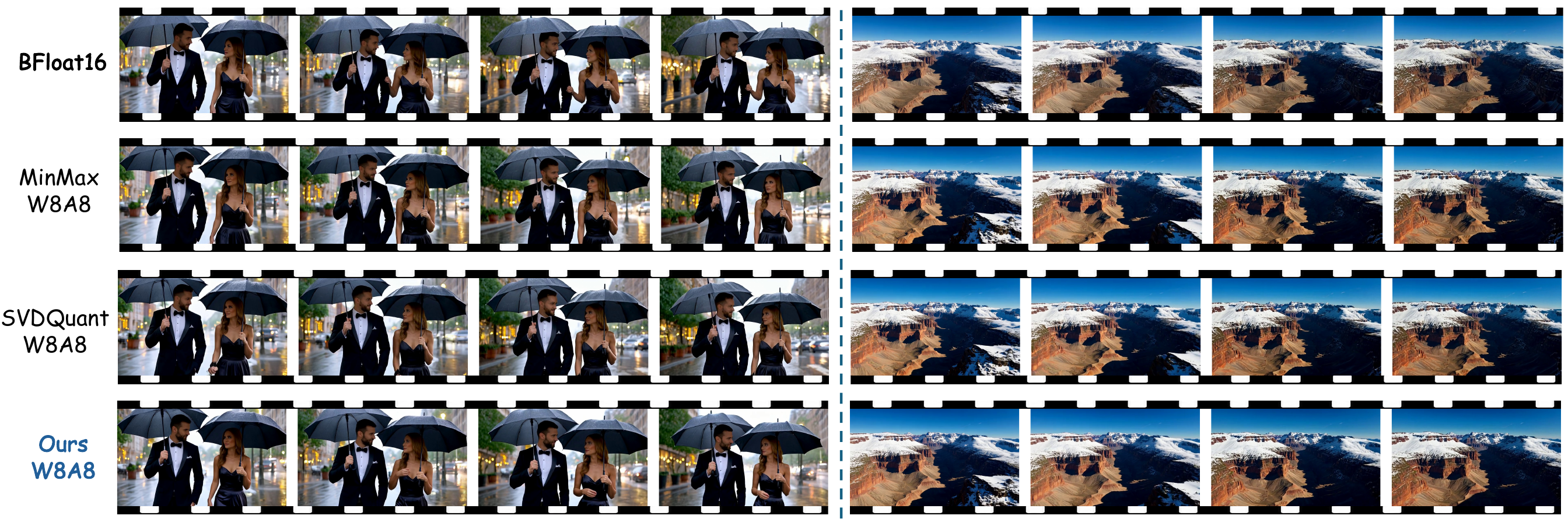}
    \caption{Visual comparison of the self-forcing model using W8A8.}
    \label{fig:appendix_visual_self_w8a8}
\end{figure*}

\begin{figure*}[htbp]
    \centering
    \includegraphics[width=\linewidth]{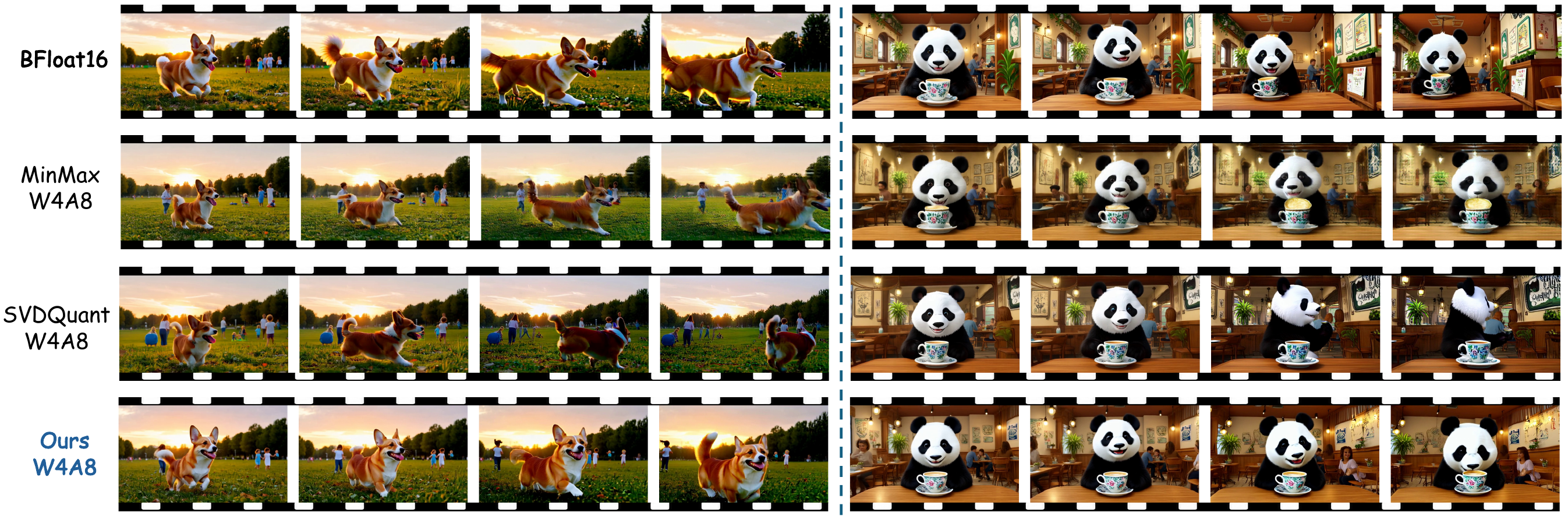}
    \caption{Visual comparison of the causal-forcing model using W4A8.}
    \label{fig:appendix_visual_causal_w4a8}
\end{figure*}

\begin{figure*}[htbp]
    \centering
    \includegraphics[width=\linewidth]{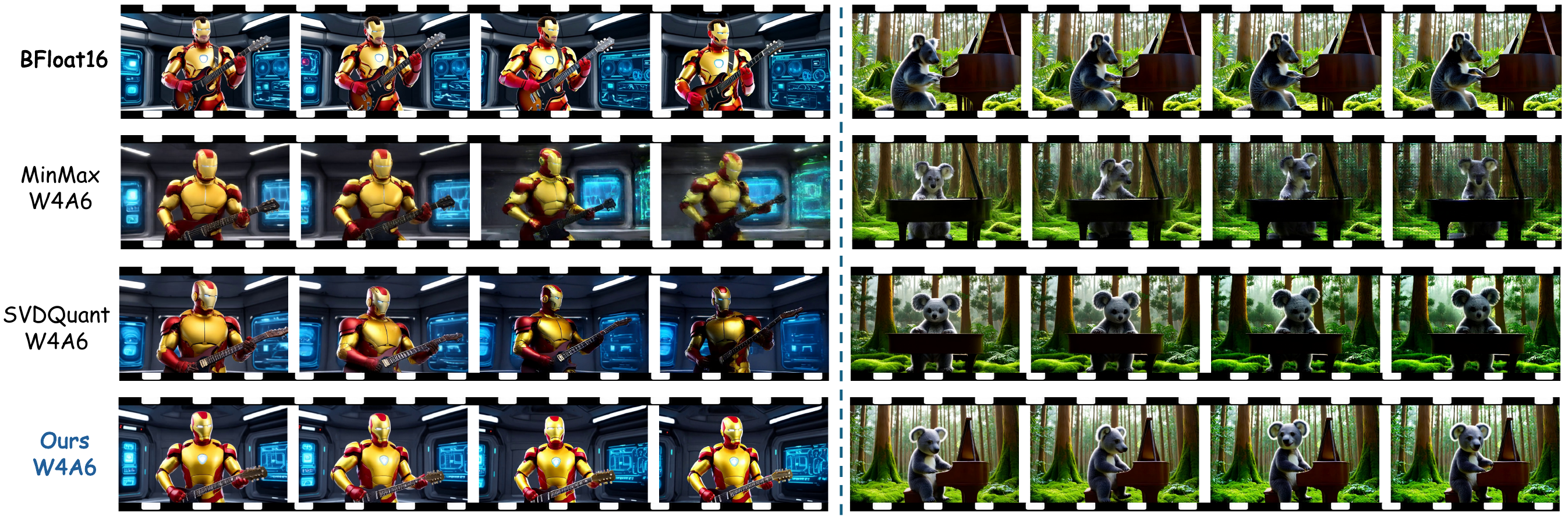}
    \caption{Visual comparison of the causal-forcing model using W4A6.}
    \label{fig:appendix_visual_causal_w4a6}
\end{figure*}

\begin{figure*}[htbp]
    \centering
    \includegraphics[width=\linewidth]{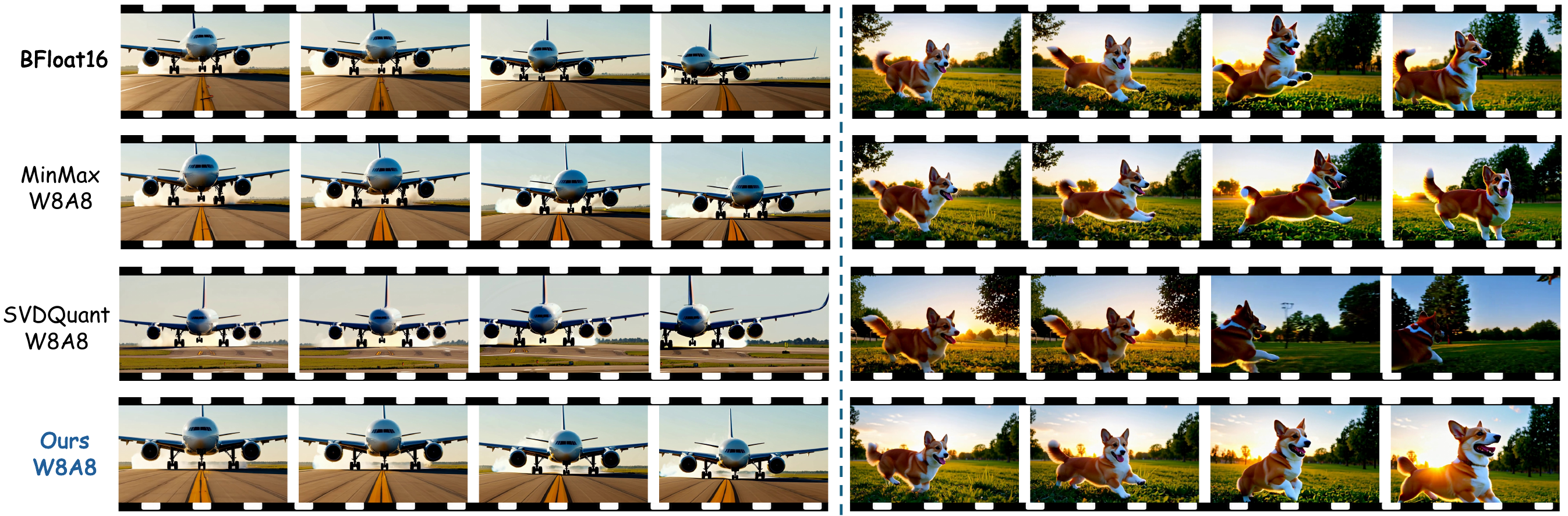}
    \caption{Visual comparison of the causal-forcing model using W8A8.}
    \label{fig:appendix_visual_causal_w8a8}
\end{figure*}

\Cref{fig:appendix_visual_self_w4a8}, \Cref{fig:appendix_visual_self_w4a6}, and \Cref{fig:appendix_visual_self_w8a8} show visual results on self-forcing using bitwidth W4A8, W4A6, and W8A8. \Cref{fig:appendix_visual_causal_w4a8}, \Cref{fig:appendix_visual_causal_w4a6}, and \Cref{fig:appendix_visual_causal_w8a8} show visual results on causal-forcing using bitwidth W4A8, W4A6, and W8A8. 
These qualitative results show that our method preserves video quality well in all bitwidth settings, while baselines induce noticeable quality degradation in low-bit W4A8 and W4A6 settings.

\section{Additional Implementation Details and Extra Cost Analysis}

Quantization Implementation. We find that the time embedding and projection modules, and the final prediction head are important to maintaining model performance. Considering their tiny size, we preserve them in BFloat16; In the reconstruction, we set the batch size to 8, the learning rates of rounding and scale to 2e-3 and 4e-5 respectively, and the total training iterations to 2000. We use the Adam optimizer with cosine annealing schedule.

Quantization Kernels and Extra Cost Analysis. We use Triton~\citep{tillet2019triton} to implement quantization kernels. We divide the quantization into two kernels. (i) Activation quantization kernel, which quantizes input float activations to INT8. (ii) INT8 GEMM and de-quantization kernel, which completes matrix multiplication of INT8 weight and INT8 activation, and also performs de-quantization based on scales. For our dual-scale quantization, we extract the outlier channels and normal channels offline, so we only need to extract the corresponding activations online (do permutation based on the split of outlier and normal weight channels), and we observe this cost is negligible. Besides, the dual-scale is only applied to part of the layers which have outliers, and remains unchanged for smooth layers. Overall, we observe a 1.30x (18.02s to 13.85s) latency speedup on NVIDIA A6000 GPU and a 1.97x (2.64GB to 1.34 GB) model size reduction, with a batch size of 2 and default inference configurations of self-forcing (4 denoising steps, 7 chunks, 21 frames, etc).

\section{Discussion about Related Outlier-handling Methods}
\label{sec:appendix_discuss}

How to address outlier issue is a long-standing question in the field of model quantization. As shown in~\Cref{eq:round_error}, outliers usually increase the quantization errors. There are several mainstream paradigms to tackle outliers. Scaling-based methods, exemplified by SmoothQuant~\citep{xiao2023smoothquant}, scale the weights and activations simultaneously in a channel-wise way, to suppress outliers in activations. However, it is not a free lunch, since it just transfer the quantization difficulty from activations to weights.
Rotation-based methods, represented by QuaRot~\citep{ashkboos2024quarot}, apply orthogonal transformations to weights and activations, aiming to obtain smoother distributions. Nevertheless, there is no theoretical guarantee that these rotated distributions are consistently favorable, and rotation can sometimes amplify certain channels in practice. Moreover, offline rotation is inapplicable in DiTs due to the usage of adaptive normalization layers~\citep{peebles2023scalable} and non-linear activation functions, while online rotation incurs notable extra overhead~\citep{lisvdquant,li2025dvd,liu2026ptq4arvg}.
Finally, low-rank branch methods, such as SVDQuant~\citep{lisvdquant}, mitigate weight outliers by absorbing them into a high-precision low-rank branch, which inevitably incurs additional computational overhead. 
Our outlier adaptive dual-scale method isolates the outliers from normal values,  thereby reducing the scaling factor for normal channels.
Guaranteed by~\Cref{eq:round_error}, the reduced scaling factor will theoretically ensure lower quantization errors. Besides, the dual-scale strategy only incurs negligible extra costs (the extraction/permutation of activations according to the split of outlier and normal weight channels), and is only applied to part of the layers which have outliers.

\section{Ablation of Minimum Constraint Alpha}

In~\Cref{eq:new_threshold}, we introduce $\alpha$ to impose a minimum outlier threshold, which helps prevent false outlier detection. Specifically, $\alpha$ should not be set too large, as this would raise the outlier threshold. Conversely, it should not be too small, as this may make the constraint ineffective and lead to false detection. We set $\alpha=1.20$ in all experiments and find it works well. Nevertheless, we further conduct an ablation study on $\alpha$, as reported in~\Cref{tab:appendix_alpha}. The results show that our method is robust over a reasonable range of $\alpha$.

\begin{table}[h]
\centering
\caption{Ablation of minimum constraint $\alpha$ using self-forcing.}
\label{tab:appendix_alpha}
\resizebox{0.5\textwidth}{!}{%
\begin{tabular}{@{}c|ccccc@{}}
\toprule
alpha & 1.10   & 1.15   & 1.20   & 1.25   & 1.30   \\ 
FVD   & 115.28 & 117.26 & 116.26 & 114.35 & 119.96 \\
LPIPS & 0.460  & 0.462  & 0.466  & 0.463  & 0.462  \\ \bottomrule
\end{tabular}
}
\end{table}

\section{Limitations}

While our method identifies an exponential-like frame-wise sensitivity, it currently leverages this property only during quantization reconstruction. This sensitivity could potentially be extended to other techniques, such as mixed-precision quantization. Besides, while we deploy our model using Triton, manually implementing CUDA kernels with dedicated optimizations might offer additional efficiency gains.


\end{document}